\documentclass[10pt,twocolumn,letterpaper]{article}

\usepackage{cvpr}              

\usepackage{graphicx}
\usepackage{amsmath}
\usepackage{amssymb}
\usepackage{booktabs}
\usepackage[accsupp]{axessibility}

\usepackage[pagebackref,breaklinks,colorlinks]{hyperref}

\usepackage[capitalize]{cleveref}
\crefname{section}{Sec.}{Secs.}
\Crefname{section}{Section}{Sections}
\Crefname{table}{Table}{Tables}
\crefname{table}{Tab.}{Tabs.}

\def\cvprPaperID{3} 
\def\confName{CVPR}
\def\confYear{2023}

\DeclareMathOperator*{\argmin}{arg\,min}

\newcommand{\sota}{\textit{state-of-the-art }}
\newcommand{\sfm}{\textit{SfM }}
\newcommand{\sfms}{\textit{SfMs}} 

\begin{document}

\title{\textit{Pointless} Global Bundle Adjustment With Relative Motions Hessians}
%
%
\author{Ewelina Rupnik\\
{\tt\small ewelina.rupnik@ign.fr}
\and 
Marc Pierrot-Deseilligny\\
{\tt\small marc.pierrot-deseilligny@ensg.eu}\\
\and
	Univ Gustave Eiffel, LASTIG, ENSG-IGN, F-94160 Saint-Mandé, France}
\maketitle
%
\begin{abstract}
Bundle adjustment (BA) is the standard way to optimise camera poses and to produce sparse representations of a scene. 
However, as the number of camera poses and features grows, refinement through bundle adjustment becomes inefficient. Inspired by global motion averaging methods, we propose a new bundle adjustment objective which does not rely on image features' reprojection errors yet maintains precision on par with classical BA. Our method averages over relative motions while implicitly incorporating the contribution of the structure in the adjustment. To that end, we weight the objective function by local hessian matrices -- a by-product of local bundle adjustments performed on relative motions (e.g., pairs or triplets) during the pose initialisation step. Such hessians are extremely rich as they encapsulate both the features' random errors and the geometric configuration between the cameras. These pieces of information propagated to the global frame help to guide the final optimisation in a more rigorous way.
   We argue that this approach is an upgraded version of the motion averaging approach and demonstrate its effectiveness on both photogrammetric datasets and computer vision benchmarks. The code is available at https://github.com/erupnik/pointlessGBA
\end{abstract} 
\section{Introduction}
\label{sec:intro}
\begin{figure}[t]
    \centering
    \includegraphics[width=0.4\textwidth]{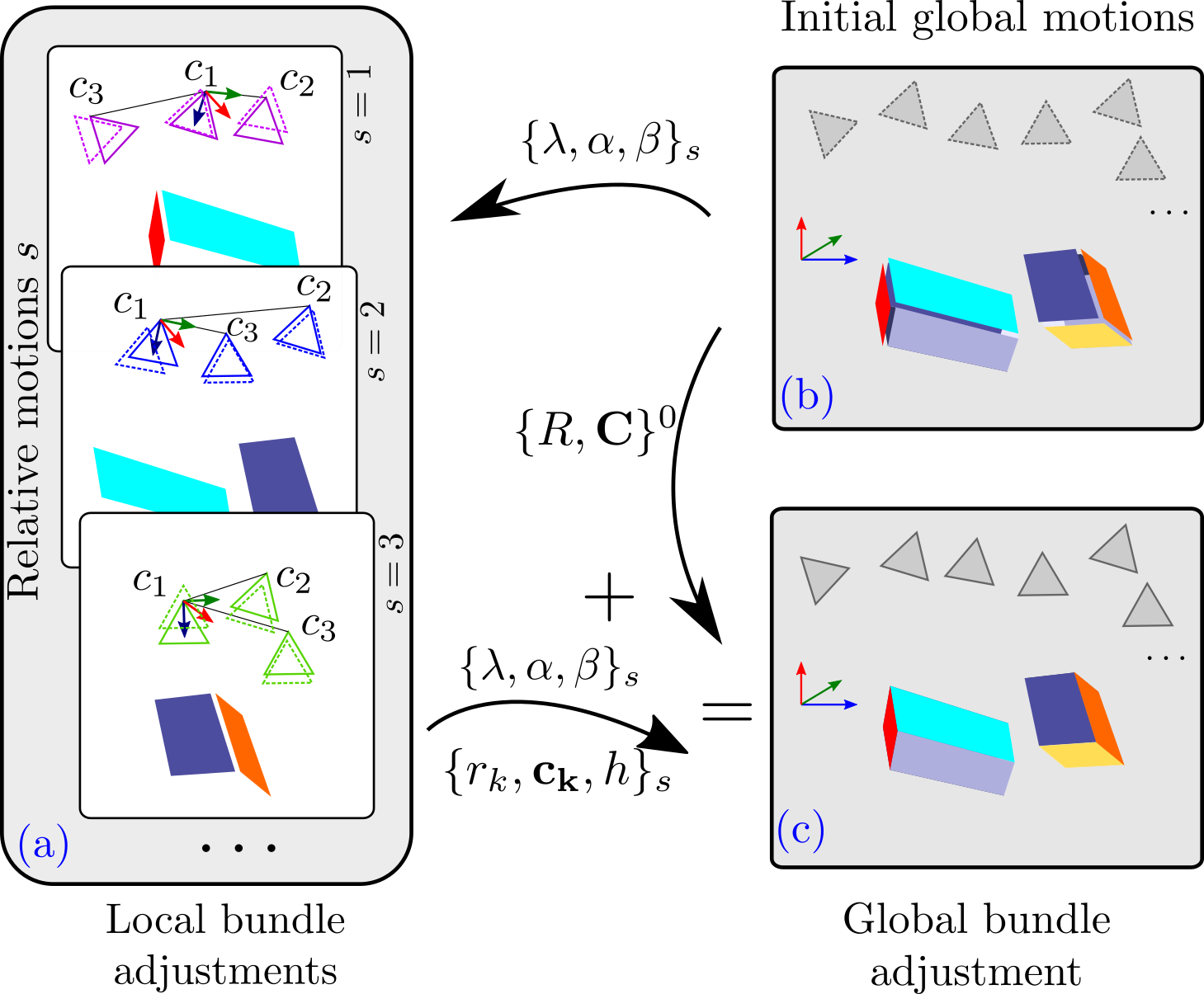}
    \caption{\textbf{\textit{Pointless} BA pipeline}. We refine global camera poses (and thus the 3D structure) in global bundle adjustment by rigorously taking into account the stochastics of the relative motions. Our inputs are $S$ relative motions $\{{r}_k,\mathbf{c}_k \}_s$~\textcolor{blue}{(a)}, their initial 3D similarity transformations $\{\lambda,\alpha,\beta \}_s$ relating them to the global frame, and initial global poses $\{R,\mathbf{C}\}^0$~\textcolor{blue}{(b)}. 
    We first run in parallel $S$ local bundle adjustments to retrieve \textit{camera reduced matrices} ${h}_s$ which encapsulate the rich stochastic information.
    We then find the optimal camera poses~\textcolor{blue}{(c)} by combining all our inputs, including the ${h}_s$ matrices. Concretely, our refinement minimises an error metric defined as the difference between the observed (\textcolor{magenta}{\textbf{--}},\textcolor{blue}{\textbf{--}},\textcolor{green}{\textbf{--}}) and predicted relative motions (\textcolor{magenta}{\textbf{- -}},\textcolor{blue}{\textbf{- -}},\textcolor{green}{\textbf{- -}})~\textcolor{blue}{(a)}, where the predictions are obtained by applying a 3D similarity to the initial global camera poses. Additionally, the error is \textit{weighted} by the ${h}_s$ matrix which virtually incorporates the effect of feature points in the adjustment. 
    In this example $k\in <1,3>$, and $S\in <1,3>$. 
    }
    \label{fig:our_pipeline}
\end{figure}
%
Photogrammetry and computer vision are nowadays widely used to produce up-to-date 2D and 3D maps of territories on a national scale as well as at the level of a city, for cultural heritage documentation, in agriculture, geology, gaming and many other domains~\cite{remondino2006image}. To generate convincing 3D representations of a scene, hundreds or thousands of images are usually involved.
More importantly, quality of the reconstructed 3D scene relies heavily on the quality of the  camera positions and rotations, the so-called camera poses. 

Our work focuses on bundle adjustement (BA)~\cite{triggs2000bundle} -- a refinement step advantageous for finding the most optimal camera poses by taking simultaneously into account {all available observations} relative to a set of images (i.e., image features, ground control points, \textit{a priori} knowledge on perspective centers or rotations, etc.). Such refinement can occur twice during an \sfm (\textit{Structure from Motion}) pipeline~\cite{schonberger2016structure}: (1) as a systematic phase to avoid error accumulation and the subsequent drift effect when incrementally building the initial solution, and (2) as a final adjustment once all images have been initialised.

As the numbers of images grow, BA routines quickly become inefficient. Solving the arising systems of equations with exact methods such as Levenberg-Marquadt implies growing space and time complexities by the second and third power in the number of BA parameters~\cite{agarwal2010bundle}. The common way to address the high computational cost is to exploit the particular structure of BA equations. The strategy known as the \textit{Schur trick} involves rearranging the equations such that the unknowns corresponding to the (few) camera parameters form an independent block, thus can be solved without intervention of 3D points. This said, for very large problems matrix rearranging and construction of  the Schur complement also becomes prohibitive~\cite{schonberger2016structure}.

To further reduce this burden, one can exploit the structure of the camera graph (i.e., viewgraph), divide a large problem in many smaller sub-problems and treat them separately, as is done in
hierarchical or hybrid \cite{toldo2015hierarchical,bhowmick2017divide,cefalu2017hierarchical} \sfms. The splitting is typically carried out via graph partitioning, then each small problem is solved independently with direct methods (i.e., space resection, F-Matrix, etc.), and aggregated in a common frame (e.g., with global or structure-less approaches, or 3D similarity transformation). This protocol is interleaved with bundle adjustments as the solution is progressively built which assures optimal results but imposes a certain processing cost. Similar in concept but different in execution is the consensus based bundle adjustment (CBA) \cite{eriksson2016consensus,mayer2019rpba}. Unlike previous approaches, CBA breaks an \textit{SfM}'s objective function into parts and solves it in a distributed way while preserving a \textit{consensus} at the break points.

The new global motion averaging~\cite{govindu2001combining,martinec2007robust} and structure-less~\cite{indelman2012incremental}  approaches to camera pose estimation both factor out the structure from the estimation problem and leverage the geometric constraints between cameras.
While this manoeuvre reduces the computation times significantly, there remains a trade off in the precision of the recovered poses. Global motion methods are thus very good at initialisating an \sfm but never considered optimal.   

\paragraph{Contributions of this paper}

Our work on bundle adjustment extends the global motion averaging methods and is presented in \cref{fig:our_pipeline}. We address their compromised precision while maintaining their computational efficiency, ultimately transforming them into optimal solutions, as opposed to being merely initialisation methods. 
We achieve this goal by indirectly incorporating information about the removed structure. More precisely, {we define our \textit{pointless}  global bundle adjustment as a function of local Hessians (i.e., the inverse of the covariance) constructed during the relative motions computation} (i.e., pairs, triplets). In doing so, the quality of the relative motions, including the random errors related to features and the correlations between camera parameters, is propagated to the global solution. This approach is similar in philosophy to \cite{rupnik2020towards} where the authors attempt to propagate the structure information per relative motion at a low cost by compressing it to 5 points. Here, in contrast, we rigorously propagate equivalent information while supplanting entirely the points from the equation. We also note that our approach is not restrained to motion averaging methods. It can be similarly adopted in any \sfm method that builds a consistent 3D structure and camera poses from many independent sub-problems. We evaluate our approach on several datasets: a typical aerial photogrammatric dataset, two computer vision benchmarks (ETH3D~\cite{Schops_2019_CVPR}, Tanks \& Temples~\cite{Knapitsch2017}), and a challenging, very long focal length terrestrial acquisition \cite{longfocdata}. Our method is compared against global motion averaging \sfm implemented in openMVG \cite{moulon2016openmvg}, incremental \sfm in MicMac \cite{rupnik2017micmac}, 5-Point bundle adjustment \cite{rupnik2020towards} and in-house implementation of the IRLS motion averaging \cite{chatterjee2013efficient}.

This paper is organized as follows. In the next section a brief review of the global motion averaging methods is given, including a discussion on robustness. In \cref{sec:method} derivation of the proposed method is outlined, followed by a description of the adjustment pipeline implementation details in \cref{sec:pipeline}. Finally, experiments are presented in \cref{sec:experiments}.

%
\section{Related work} 
\label{sec:relatedwork}

\paragraph{Global motion averaging}
Motion averaging methods use elementary relative motions, typically pairs or triplets of images, to resolve the camera poses in a global and fast manner. Because the poses are computed all at once, motion averaging surmounts the pitfall of incremental methods \cite{schonberger2016structure} where errors accumulate all along the initialisation step, and lead to trajectory drift. However, such methods give rise to new challenges. First, by relying exclusively on pairs or triplets of images, motion averaging methods ultimately renounce higher observation redundancy (i.e., long feature tracks), which we know negatively impacts both the camera pose estimation robustness and precision \cite{lindenberger2021pixel}. Second, once the relative poses computed, the structure used in the calculation is discarded, and all relative relationships, whether derived from erroneous observations or not, are treated equally. 

As a result, there have been many works addressing the precision as well as the mechanisms of handling low quality and outlier relative relationships in motion averaging. For instance,  \cite{govindu2006robustness} proposed sampling random spanning trees and RANSAC on the pose viewgraph (i.e., a graph where the nodes and edges represent the images and relative relationships), while \cite{zach2010disambiguating,sweeney2015optimizing} explored the viewgraph's structure to prune inconsistent loops or optimise the initial relative constraints.  {Moulon \etal} \cite{moulon2013global} leverage the trifocal tensor to strengthen the relative translation retrieval. Instead, \textit{1DSfM} \cite{wilson2014robust} casts translations as 1D problems and recovers inconsistencies through 1D graph ordering of pairwise constraints. A complete two-stage robust pipeline was introduced in \cite{crandall2011discrete}. The authors embed the cameras relative relationships and 3D points within a Markov Random Field graph, then simultaneously solve for initial camera poses using discrete belief propagation. The rotations parameterised by a set of discrete 3D rotations provide only a coarse result, which serves to eliminate outliers and initialise the subsequent continuous optimisation. 
 
Others suggest to build-in the robustness in the estimation step itself. Arrigoni \etal  \cite{arrigoni2014robust} represents the rotation averaging as a matrix decomposition problem. A measurement matrix decomposed into a sum of low-rank and sparse terms naturally groups the gross errors in the latter. Having identified the gross errors, they participate in a modified $l_2$ rotation averaging that follows, with minimal impact on the output. Nevertheless, storing all relative motions in the measurement matrix might turn prohibitive for very large scale \sfms. Instead of resorting to the non-robust $l_2$ rotation distance averaging, {Hartley \etal} \cite{hartley2011l1} rigorously average rotations in the orthogonal $SO(3)$ group through application of the $l_1$ Weiszfeld algorithm. Such formulation is equivalent of computing a geometric median over multiple rotations, and its major merit is its simplicity. To its disadvantage, the one-by-one rotation update makes it a slow convergence optimisation \cite{chatterjee2017robust}. The golden standard for robust motion averaging in the presence of outliers is unarguably the \textit{iteratively reweighted least squares} (IRLS) introduced in \cite{chatterjee2013efficient,chatterjee2017robust}. Given a set of reliable initial estimates of the global rotations (e.g., obtained with robust $l_1$ optimisation), IRLS simultaneously finds their optimal values through iterative regression. The influence of individual errors on the solution is governed by a suitable loss function. IRLS demonstrated superior performance with respect to the \sota in speed and accuracy.
 

Unlike the \sota approaches which discard entirely any information related to feature points from the global averaging, our pipeline retains and conveys the features in a compact form via local hessians. Our local hessians propagated to the global frame rigorously guide the global camera pose refinement. Outliers are handled implicitly by a robust cost function, however, we assume that the majority of gross errors has been removed prior to the adjustment.
  

%
%

\paragraph{Exploiting hessian matrices.}
The hessian matrix (or its inverse -- the covariance) resulting from a bundle adjustment encapsulate information about random observation errors, and inter-dependencies between estimated parameters, i.e., in our case the cameras and 3D points.
These information-rich matrices have been long used in photogrammetry for theoretical accuracy analyses. For instance, the \textit{a posteriori} retrieved variances and co-variances have been used  (i)~as a quality measure of 3D intersections \cite{fraser1982optimization,mayer2019rpba}, (ii) as a tool to design optimal imaging network \cite{fraser1984network} or for next-best view selection \cite{haner2012covariance}, (iii) to analyse correlation between camera intrinsic and extrinsic parameters \cite{zhou2018gnss}, as well as (iv) in airborne laser strip adjustment when GNSS/IMU trajectory is not available~\cite{ressl2012applying}. Other common uses of the covariances include \textit{Kalman filtering} in recursive pose estimations or visual SLAM \cite{davison2007monoslam}. There, each new camera pose predictions are made from a product of covariance-weighted current state and available new measurements. To the best of our knowledge, this paper is the first to exploit hessian/covariance matrices in global motion averaging.



 
\section{Global optimisation with local Hessians} 
\label{sec:method}

\paragraph{Problem formulation}
Building a global orientation of a block of images involves two steps: recovery of the initial global orientation of all images through incremental, global or hybrid \textit{SfM}; followed by a final bundle adjustment that refines simultaneously all poses and the 3D structure.
Our goal is to refine initial poses $ \{{R},\mathbf{C} \}_j^0$ of a number of images where ${R}$ is a rotation matrix and $\mathbf{C}$ is a perspective center defined in the global reference frame. However, unlike in the classical BA that minimises the point's re-projection errors, our \textit{pointless}  BA's objective function relies exclusively on three ingredients (cf. \cref{fig:our_pipeline}): 
\begin{itemize}
    \item  the relative motions, 
    \item per-motion hessian matrices -- a by product of the relative motions' estimation (i.e., the local bundle adjustment), and 
    \item the initial 3D similarity transformations relating the global and the local frame of the relative motion.
\end{itemize}

Differently to the standard IRLS approach which considers all relative motions as static, our motions come with unique uncertainty signatures contained in the hessian matrices. Those are subsequently integrated in the global cost function minimising over all camera poses.  

For the sake of completeness of this derivation, in the coming section we lay out the local bundle adjustment step and hessian retrieval. We then follow up with global to local frame propagation and the derivation of our \textit{pointless} global bundle adjustment cost function.
\paragraph{Local bundle adjustment.}
For every relative motion composed of $N$ views and $M$ features, we can write the cost function expressed in local frame of the relative motion as:
\begin{equation}
\begin{split}
    E_{BA}^l = \sum_{k=0}^{N} \sum_{i=0}^{M} \left( F( \mathbf{p}_i) \right)^2 \\
    = \sum_{k=0}^{N} \sum_{i=0}^{M} \rho_{ki} \left(  f( \mathbf{p}_i ) - \mathbf{o}_{ki} \right)^2,
\end{split}\label{eq:costgen}
\end{equation}
where $\mathbf{o}_{ik}$ are the observations corresponding to image features in $k^{th}$ view, and $\mathbf{p}_i$ are their respective 3D coordinates expressed in the local frame of the relative motion. The function $f(\cdot)$ relates a 3D point $\mathbf{p}_i$ with its predicted image observation $\mathbf{\bar{o}}_{ki}$ and follows the known projection function with $\mathcal{J}$ as the intrinsic parameters, and $\{ {r}_k, \mathbf{c}_k \}$ as the extrinsic parameters: $f(\mathbf{p}_i) = \mathbf{\bar{o}}_{ki} = \mathcal{J}_{k}  \left( \pi_{k} \left( {r}_k \left( \mathbf{p}_i - \mathbf{c}_k \right) \right) \right)$. The loss function $\rho$ reduces the impact of outliers on the solution.

By minimising the quadratic form in \cref{eq:costgen} we obtain $\delta \mathbf{x}$ updates to all unknowns (i.e., extrinsic parameters and the 3D coordinates of feature points):
\begin{equation}
\begin{split}
     \delta \mathbf{x}^* = \argmin_{\delta \mathbf{x}} ( J \delta \mathbf{x} + F_0)^2 =\\
      \argmin_{\delta \mathbf{x}} \left( \delta \mathbf{x}^T \underbrace{J^T J}_{H} \delta \mathbf{x} + \underbrace{2 F_0^T J}_{G} \delta \mathbf{x} + F_0^2 \right) \\
      \equiv
      -H^{-1} \cdot G,
\end{split}\label{eq:costQuad}
\end{equation}
where $J$ is a $\left(2MN \times 6N+3M\right)$ Jacobian matrix, $H$ and $G$ are the hessian (aka the normal equations) and the gradient of the cost function, $F_0$ is the value of the cost evaluated at current estimate of the unknowns, and $\delta \mathbf{x}$ is the difference between the current $\mathbf{x}$ and initial $\mathbf{x}_0$ estimate of the unknowns. 

The hessian matrix in \cref{eq:costQuad} describes all unknowns while we are only interested in the unknowns corresponding to the extrinsic parameters. Thus, we re-write it with the help of the \textit{Schur complement}, and note ${h}$ the $6N \times 6N$  \textit{camera reduced matrix}.  We then transcribe the cost in \cref{eq:costQuad} to a cost relying only on the relative camera extrinsics:

\begin{equation}
\begin{split}
    \delta \mathbf{x}^* = \argmin_{\delta \mathbf{x}} \left( {\delta \mathbf{x}}^T \cdot h \cdot {\delta \mathbf{x}} +  {g}^T {\delta \mathbf{x}} + \mathbf{m} \right)~.
\end{split}\label{eq:costquadraticSchur} 
\end{equation}

\paragraph{Global to local frame propagation.}
Note that the local extrinsic parameters  $\{\mathbf{c}_k,r_k \}$  are related to their global equivalents  $\{\mathbf{C},R \}$  by a 3D similarity transformation  ${d}$:  

\begin{equation}
   \mathbf{x}_{k} =  \{ \overbrace{ \lambda \cdot \alpha \cdot \mathbf{C} + \mathbf{\beta}}^\text{$\mathbf{c}_{k}$} , \overbrace{\alpha \cdot R}^\text{$r_{k}$} \} =  {d} \left( \{\mathbf{C},R \} \right)~,
   \label{eq:similarity}
\end{equation}
where $\mathbf{x}_{k}$ is a $6 \times 1$ vector of the local extrinsics of $k^{th}$ view within some relative motion; $\lambda$, $\alpha$ and $\mathbf{\beta}$ are the scale factor, $3\times3$ rotation matrix and $3 \times 1$ translation vector between the local and global frames. By injecting \cref{eq:similarity} in \cref{eq:costquadraticSchur} we can express our cost function in terms of the global camera extrinsic parameters. Observe that optimising the cost written in this way will change the initial global poses by rigorously taking into account the stochastic properties of the parameters computed in the relative frame and encapsulated within the camera reduced matrix ${h}$.

\paragraph{\textit{Pointless}  global bundle adjustment.}
Our objective is to compute refined camera extrinsics by integrating three pieces of information in a global bundle adjustment: relative motions, their local hessians, and the transformation relating local and global frames. For convenience, we transform the quadratic cost in \cref{eq:costquadraticSchur} to a sum of linear terms which can then be readily used in any least squares solver. To do that, we decompose the small hessian into $6N \times 6N$ matrix $V$ of eigenvectors and the corresponding eigenvalues matrix $D$. Furthermore, we integrate the global poses in the cost function by predicting the current estimate of the relative motion from its corresponding current global values (see \cref{eq:similarity} and \cref{fig:our_pipeline}). With this, our global bundle adjustment cost function defined over $S$ relative motions takes the following form:
\begin{equation}
    \begin{split}
        E_{BA}^g = \sum_{s=0}^S E_{s}^g = \sum_{s=0}^S {\delta \mathbf{x}_s}^T \cdot h_s \cdot {\delta \mathbf{x}_s}  \\ 
        = \sum_{s=0}^S {\delta \mathbf{x}_s}^T \cdot V_s^T D_s^2 V_s \cdot {\delta \mathbf{x}_s} \\   
        = \sum_{s=0}^S \left(  D_s \left( V_s \cdot \delta \mathbf{x}_s  \right)  \right)^2 \\ 
        = \sum_{s=0}^S \left(  D_s \left( V_s\cdot {d}(\mathbf{X}) - V_s \cdot \mathbf{x}_{0s} \right)  \right)^2 ,
    \end{split}\label{eq:globalBA}
\end{equation}
where the relative motion parameters $\mathbf{x}_{0}$ are the \textit{observations} in the adjustment, while the global camera poses $\mathbf{X}$, and the 3D similarity parameters $\{ \lambda, \alpha, \beta \}$ within ${d}$ are the \textit{unknowns} with known initial values. Every relative motion adds a $6N \times 1$ observation vector to the global cost, and the number of observations accumulated over all motions equals $6 N S$. We omit the gradient $\mathbf{g}$ and the constant $\mathbf{m}$ terms because their values are cancelled in the preceding relative motion bundle adjustment. 
\begin{figure*}[h!]
    \centering
    \includegraphics[width=1.00\textwidth]{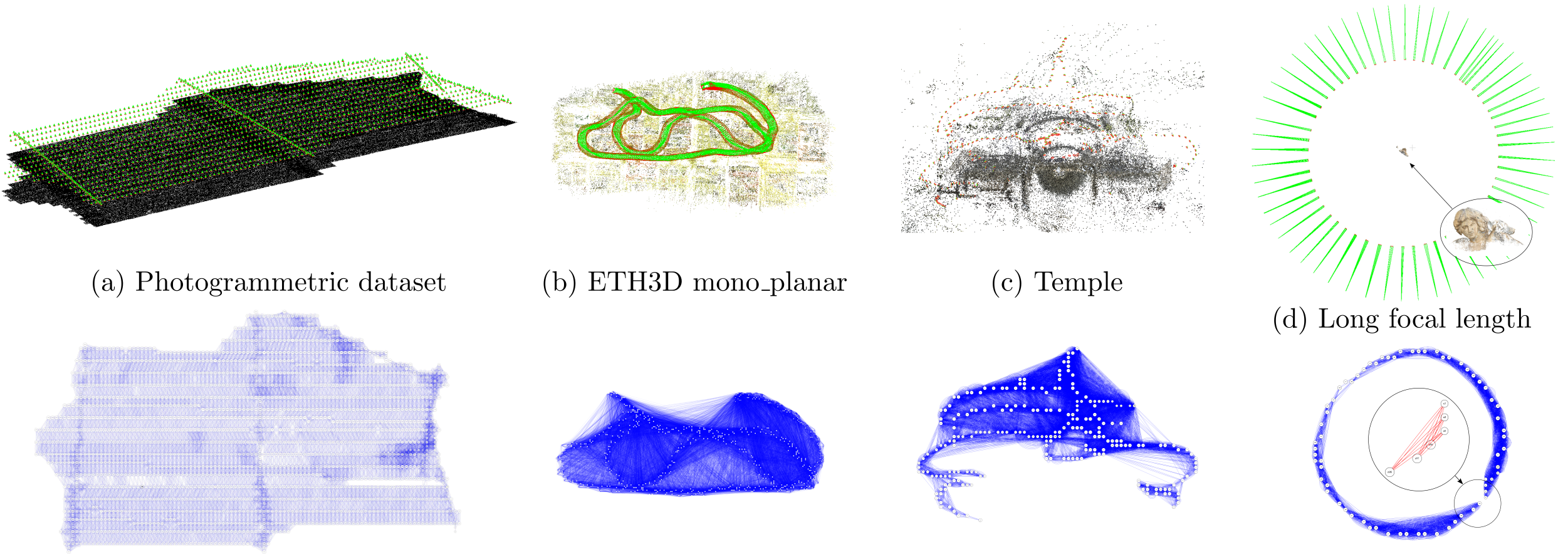}
    \caption{\textbf{Datasets.} We test our method on a classical photogrammetric aerial acquisition, two computer vision benchmarks (ETH3D, Temple) and a challenging long focal length scenario. Top: Camera poses (in green and red) and sparse 3D structure. Bottom: Triplet graphs where the blue edges correspond to known relative motions. In (d) during testing only blue edges are exploited (i.e., no loop), while in evaluation the trajectory's drift is computed using feature points common to images linked by the red edges.}
    \label{fig:all_datasets}
\end{figure*}

\paragraph{Complete adjustment pipeline.}\label{sec:pipeline}
Taking all the ingredients into account, the full pipeline involves the following steps: 
\begin{enumerate}
    \item features extraction (e.g., SIFT \cite{lowe1999object}), 
    \item generation of observations, including the relative motions and the initial global solution,
    \item per-motion local bundle adjustments, and
    \item propagation and refinement in global bundle adjustment.
\end{enumerate}

We rely on MicMac solution \cite{rupnik2017micmac} for steps 1--2, and limit the relative motions set to three-view relationships (i.e., triplets), thus $N=3$. This choice is justified by the fact that triplets (i)  provide additional redundancy hence are more reliable than pairs, and (ii) they are easy to compute thanks to the powerful modern feature extractors. To obtain the hessian matrices we run, in parallel, single-iteration local bundle adjustments with triplet poses and SIFT features from steps 1--2 as inputs. Note that steps 2 and 3 are typically seamlessly performed in a single step. We rely on a third-party solution for relative motion thus we separate them in two. Finally, the outputs from steps 2 are 3 are used to simultaneously refine all initial global poses.
%
%

\section{Experiments}
\label{sec:experiments}
\subsection{Implementation details}

\paragraph{Rotation parameterisation.}
Rotations in 3D Euclidean space form a special orthonormal group $SO(3)$. Optimising rotations without taking extra precautions might destroy this property. Among the common parameterisations that conserve the matrix orthogonality are the Lie algebra, angle-axis representation or quaternions \cite{hartley2013rotation}. We describe the rotations as a product of the known initial rotation $R_0$ and an unknown skew-symmetric small rotation $\omega_\times$: $\hat{R} = R_0 \left( I + \omega_\times \right)$. We enforce the orthogonality of the final rotation by mapping it to the closest rotation with SVD \cite{martinec2007robust}. 
The small rotation matrix is initially set to zero and optimised during the adjustment. 

\paragraph{Local and global bundle adjustments.}
We run single-iteration local bundle adjustment per each triplet following the cost defined in \cref{eq:costgen}. Dense Shur solver of Ceres library \cite{Agarwal_Ceres_Solver_2022} is used for optimisation. The inputs are: a triplet of images with their initial relative poses and image features. Our cost function is weighted by a Huber loss, and an attenuation loss $\gamma$. The first minimises the influence of the outliers, while the latter harmonises the triplets between them in terms of the number of feature points. We want to avoid penalising triplets with many features which might naturally lead to larger hessian values. To that end, we weight each image feature observation by $\gamma$ which {simulates} an equal number of observations for everyone: $\gamma = \frac{M \cdot Q}{M + Q}$ where $Q$ is the fictitious number of points, and $M$ is the input number (in our experiments $Q=10$). To compute the inverse of the local hessians one must fix the gauge ambiguity. This can be done in many ways, for instance by fixing the pose of the first camera and the base between the first and second camera, or by considering all camera extrinsics as observed. In our experiments we choose the latter. Triplets with less than 30 image features are ignored in the processing. 
  
In the global adjustment, we accumulate observations corresponding to all triplets in the triplet graph following the \cref{eq:globalBA} and  solve it using sparse Shur solver in Ceres~\cite{Agarwal_Ceres_Solver_2022}. In analogy to IRLS we weigh the observations by the residual fitting error and apply the Huber loss.

 \begin{table*}[th!]
 \caption{\textbf{Reprojection errors}. We evaluate the precision of Our$_{BA}$ and compare it with competitive methods. $\sigma_{init}$ and $\sigma_{final}$ are the initial and final reprojection errors, $Aver_{BA}$ corresponds to our implementation of IRLS \cite{chatterjee2013efficient}, while $\#param$ is the number of unknowns constituting the BA problem ($k \equiv \times 10^3$). The difference between (b) and (c) is in the size of the triplet graph, the latter being filtered to contain $\approx10\%$ of the initial count of relative motions. High residuals in (d) are due to the presence of outliers among the features. All methods except for openMVG were initialised with the same approximate global poses. Ours$_{BA}$ performs as good as the BAs within incremental \sfms~and the light 5-Pts$_{BA}$; Aver$_{BA}$ performs least good.}
     \label{tab:results_all}
    \begin{subtable}{0.33\textwidth}
        \centering
        \caption{\textbf{Photogrammetric dataset}  }
          \label{tab:photo}
          \begin{tabular}{@{}lccc@{}}
            \toprule
            \textbf{Method} & $\sigma_{init}$ & $\sigma_{final}$ & $\#$params  \\
            \toprule
            MicMac$_{BA}$ & 29.79 & 0.27 & 5,545k \\
            oMVG$_{GBA}$ & -- & 0.27 & -- \\ \midrule
            5-Pts$_{BA}$ &    & \textbf{0.28} & 799k \\
            Ours$_{BA}$ & 29.79  & \textbf{0.28} & 135k \\
            Aver$_{BA}$ &    & 2.65 & 135k \\
            \bottomrule
          \end{tabular}
    \end{subtable}
    \hfill
    \begin{subtable}{0.21\textwidth}
        \centering
        \caption{\textbf{ETH3D planar\_mono} }
        \label{tab:ethplanar}
        \begin{tabular}{@{}ccc@{}}
        \toprule
          $\sigma_{init}$ & $\sigma_{final}$ & $\#$params \\
        \toprule
          14.69 & 0.56  & 1,388k \\
           -- & 0.57 & --\\ \midrule
             & \textbf{0.56} &  5,136k \\
           14.69  & \textbf{0.56}  & 2,372k \\
             & 0.87 &  2,372k\\
        \bottomrule
      \end{tabular}
     \end{subtable}
     \hfill
    \begin{subtable}{0.18\textwidth}
        \centering
        \caption{\textbf{ETH3D planar\_mono} }
      \label{tab:ethplanar_small}
        \begin{tabular}{@{}cc@{}}
        \toprule
           $\sigma_{final}$ & $\#$params \\
        \toprule
            0.56  & 1,388k \\
            0.57 & --\\ \midrule
              \textbf{0.56} &  518k \\
              0.61 &  244k  \\
              1.93   &  244k \\
        \bottomrule
      \end{tabular}
     \end{subtable}
     \hfill
    \begin{subtable}{0.22\textwidth}
        \centering
        \caption{\textbf{Temple} }
       \label{tab:temple}
        \begin{tabular}{@{}cccc@{}}
        \toprule
          $\sigma_{init}$ & $\sigma_{final}$ & $\#$params   \\
        \toprule
          15.92  & 3.66  & 224k \\ 
          -- & 4.94 & -- \\ \midrule %
           15.92  & \textbf{3.68} &  110k \\ 
    
           15.92  &  3.72  &  49k \\
          15.92   & 6.77 &  49k \\
        \bottomrule
      \end{tabular}
     \end{subtable}  
\end{table*}

\begin{table}[h]
\caption{\textbf{Loop-closure error}. For the long focal length dataset we evaluate the precision of our method and compare it with competitive methods using the loop closure metric. This metric refers to the pixel reprojection error computed on features common to images linked by red edges in \cref{fig:all_datasets}(d). $\#params$ refers to the size of the BA problem ($k \equiv \times 10^3$). In the $REF_{BA}$ we impose the closed loop and run bundle adjustment in MicMac, therefore we consider this result as our reference. Thanks to the rigorous propagation of the relative motions' stochastics, Our$_{BA}$ performs best among the \textit{fast} BAs (5-Pts$_{BA}$, Aver$_{BA}$), and almost as good as the best performing point-based BA in MicMac.}
  \label{tab:longF}
  \centering
  \captionsetup[subtable]{labelformat=empty}
\begin{subtable}{0.5\textwidth}
        \centering
        \caption{\textbf{Long focal length dataset} }  
       \begin{tabular}{@{}lccc@{}}
    \toprule
    \textbf{Method} & $err_{loop-c}$  & $\#$params \\
    \toprule  
    REF$_{BA}$ & 0.91 & 7,523k \\\hline 
    MicMac$_{BA}$ & 3.44 &  2,283k  \\  
    openMVG$_{GBA}$ & 31.08 &  -- \\ \midrule 
    5-Pts$_{BA}$ &  48.11  & 19k \\ 
    Ours$_{BA}$ &  \textbf{4.10}   &  9k   \\
    Aver$_{BA}$ &  $>$200  &  9k    \\
    \bottomrule
  \end{tabular}
\end{subtable}  
\end{table}

\subsection{Evaluation}
\paragraph{Datasets.} To evaluate our method we look at four datasets (see \cref{fig:all_datasets}): 
\begin{itemize}
    \item \textbf{Photogrammetric dataset} -- a typical photogrammetric acquisition with a 80/60\% along- and across-track overlap composed of 2000 calibrated images over a sub-urban taken with the UltraCAM Eagle (26460x17004pix, F=120mm).  
    \item \textbf{ETH3D mono\_planar} \cite{Schops_2019_CVPR} -- a SLAM benchmark, a highly overlapping video acquisition of a flat surface consisting of 630 calibrated images (739x458pix, F=726pix).
    \item \textbf{Temple} \cite{Knapitsch2017} -- a 3D reconstruction benchmark Tanks \& Temple, 282 calibrated images of a temple (1920x1080pix, F=1163pix)
    \item \textbf{Long focal length} \cite{longfocdata} -- a challenging very long focal length acquisition composed of 93 calibrated images taken around a sculpture (5616x3744pix, F=1000mm)
\end{itemize}

\paragraph{Comparisons with existing methods}
We compare our method against the bundle adjustments within the incremental \sfm in MicMac \cite{rupnik2017micmac} and the global \sfm in openMVG \cite{moulon2016openmvg}, the 5-Point BA \cite{rupnik2020towards}, and our own implementation of IRLS motion averaging \cite{chatterjee2013efficient}. 
\paragraph{Metrics.}
As our bundle adjustment objective function implicitly minimises the features reprojection error (also true for BA implementations of the \sfms~we test against), we decide to use that metric as our only evaluation measure. Comparing absolute pose accuracies would involve choosing a reference pose estimation algorithm which is known to induce a bias on the evaluation itself \cite{brachmann2021limits}. 

Moreover, in the long focal length dataset we benefit from the acquisition geometry forming a closed-loop to evaluate the trajectory's drift. During BA, the connections between the first  and last few images of the acquisition are removed (i.e., no features in common and no relative relationships, see \cref{fig:all_datasets}(d)). During evaluation, for a perfectly recovered trajectory, reprojection errors computed on features common to the beginning and the end of the acquisition should be close to zero. Nevertheless,  pose errors accumulated along the trajectory incur a trajectory drift resulting in compromised precisions (see \cref{tab:longF}).

\begin{figure}[ht!]
  \centering
  \begin{subfigure}{0.52\linewidth}
    \includegraphics[trim=50 25 40 0, clip,width=1.0\linewidth]{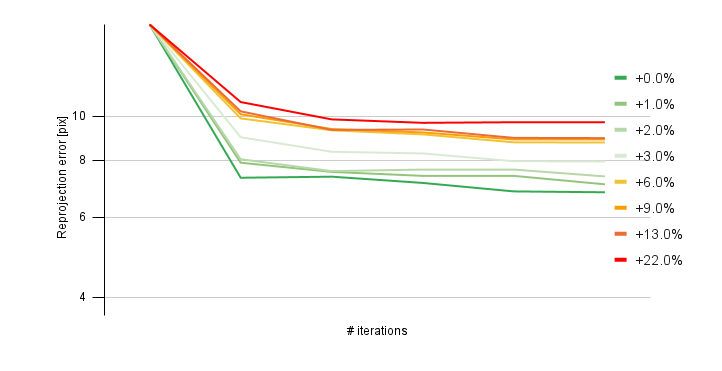}
    \caption{Aver$_{BA}$}
    \label{fig:conv_out_averBA}
  \end{subfigure}
  \hfill
  \begin{subfigure}{0.47\linewidth}
     \includegraphics[width=1.0\linewidth]{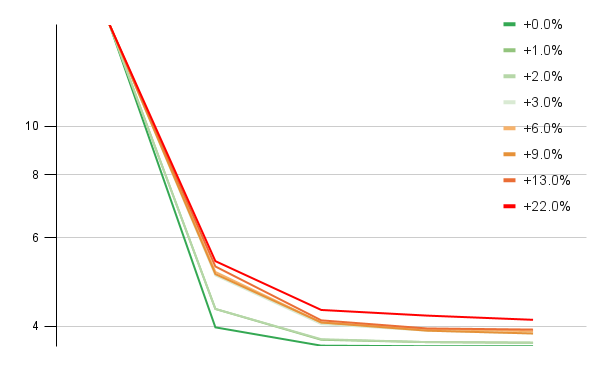}
    \caption{Ours$_{BA}$}
    \label{fig:conv_out_ours}
  \end{subfigure} 
  \caption{\textbf{Sensitivity to outliers experiment}. We infuse between 0 and 22$\%$ of outliers within the relative rotations, and observe their impact on the final reprojection errors (expressed in logscale). As the portion of outliers grows the metrics deteriorate in all cases, however, Our$_{BA}$ detoriorates at a lower pace. The $+$ signifies that outliers are added to the initial triplet graph, i.e., the accumulated ratio of outliers might be slightly higher. Sensitivity tests are performed on Temple benchmark.}
  \label{fig:convergence_outliers}
\end{figure}

To asses the sensitivity of our method to outliers we randomly infuse the relative rotations with outliers as observe their effect on the reprojection error across bundle adjustment's iterations, as shown in \cref{fig:convergence_outliers}.

The MicMac and openMVG \sfms~are complete pipelines and singleing out the runtime contribution of just the BA step is not straightforward. For that reason, we use the number of parameters per problem and the convergence rate as proxy for runtime.


\begin{figure*}[ht!]
  \centering
  \begin{subfigure}{0.23\linewidth}
    \includegraphics[width=1.0\linewidth]{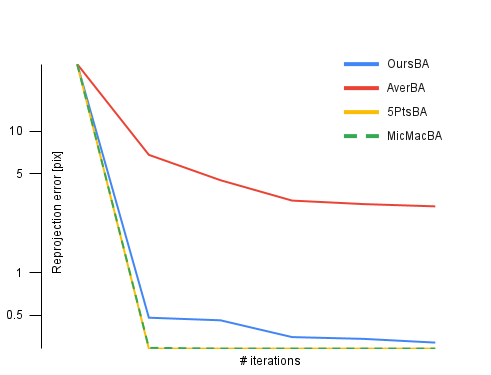}
    \caption{Photogrammetric d.}
    \label{fig:short-a}
  \end{subfigure}
  \hfill
  \begin{subfigure}{0.23\linewidth}
     \includegraphics[width=1.0\linewidth]{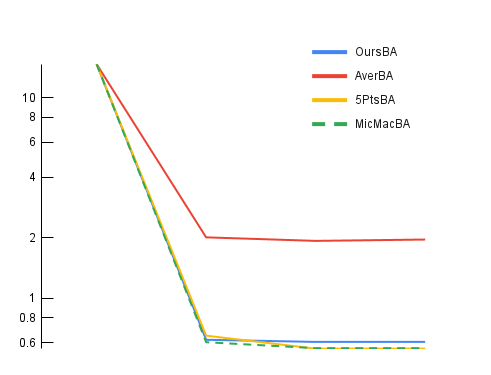}
    \caption{ETH3D mono\_planar}
    \label{fig:short-b}
  \end{subfigure}
  \hfill
   \begin{subfigure}{0.23\linewidth}
    \includegraphics[width=1.0\linewidth]{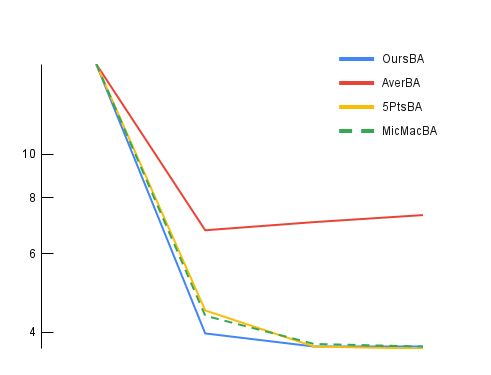}
    \caption{Temple}
    \label{fig:temple_convergence}
  \end{subfigure}
  \hfill
  \begin{subfigure}{0.23\linewidth}
    \includegraphics[width=1.0\linewidth]{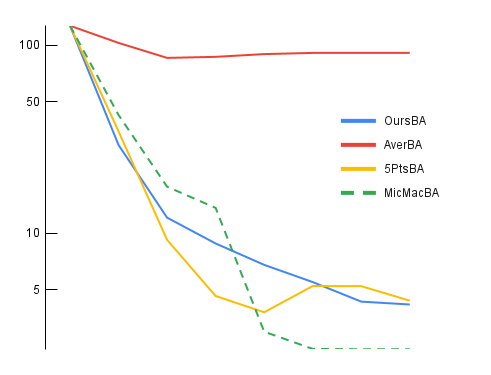}
    \caption{Long focal length}
    \label{fig:longF_convergence}
  \end{subfigure}
  \caption{\textbf{Convergence experiment}. We evaluate the rate of convergence for all of the tested methods. Our method (Ours$_{BA}$) minimizes at a rate comparable to point-based BA in MicMac and the 5-Pts$_{BA}$ across all datasets, while the version of IRLS motion averaging (Aver$_{BA}$) performs worst. Note that Our$_{BA}$ is effectively the lightest among the best-converging methods (MicMac$_{BA}$, 5-Pts$_{BA}$) because it engages much less unknowns (see \cref{tab:results_all}).  Reprojection errors are expressed in logscale.}
  \label{fig:convergence}
\end{figure*}
\subsection{Results and discussion}
Feature reprojection errors on the Photogrammetric dataset, ETH3D planar\_mono and Temple benchmarks are given in \cref{tab:results_all}, while the loop closure error on the Long focal length dataset is shown in \cref{tab:longF}. 

In terms of precision, our \textit{pointless} BA performs as good as the classical BAs and the 5-Point BA. It significantly outperforms the IRLS averaging (i.e., $Aver_{BA}$). This tendency repeats across all datasets. The trajectory loop closure error in the challenging Long focal length dataset reveals the superiority of our \textit{pointless} BA against the 5-Point BA. It highlights the power of the hessian propagation which, by bringing the stochastics of the local bundle adjustment into the global adjustment, prevents large trajectory drifts.

We reduce the size of the BA problem by at least a factor of 4 with respect to the standard BA, and up to 40 times for the Photogrammetric dataset (5,545k vs 135k unknowns). This is thanks to the controlled acquisition pattern and the resulting optimality of the dataset's viewgraph contaning a limited number of redundant triplets. Compared to 5-Point BA, we halve the number of parameters. One can safely assume that reducing the triplet graphs for other datasets would proportionally increase their reduction factors. 

As shown in \cref{fig:convergence} all tested methods except the Aver$_{BA}$ follow similar convergence rates, yet Ours$_{BA}$ with much fewer unknowns is the lightest among the best-converging. Finally, faced with outliers our hessian BA, weighted by the fitting residual error and the Huber loss function, shows only marginal deterioration of the reprojection metric (see \cref{fig:convergence_outliers}).

\paragraph{Inclusion of ground control points.} Although not presented in this study, our BA can be easily extended to include ground control points (i.e., GCPs or landmarks). To that end, the initial global solution is first transferred to the coordinate frame of the GCPs (i.e., the new global frame), and the initial 3D similarity transformations are changed correspondingly. Then, for each relative motion where a GCP is seen in at least two images, the global BA in \cref{eq:globalBA} is extended to include the GCP's residual: $ \mathbf{r}_{GCP} = \rho_{GCP}|| \mathbf{P}_{GCP} - d^{-1} \left( \mathbf{p}_{GCP} \right) ||^2$, where $\mathbf{P}_{GCP}$ and $\mathbf{p}_{GCP}$ are the GCP's 3D coordinates in global and local frames, $d^{-1}$ is the inverse 3D similarity transformation moving from the local to global frame from \cref{eq:similarity}, and $\rho_{GCP}$ is an appropriate weighting function. 

\paragraph{Self-calibrating bundle adjustment.} Our method assumes calibrated cameras with precisely known intrinsic parameters, but it could be extended to self-calibration. We lay out this extension below, stipulating that we have not conducted experiments proving its practicality or effectiveness. 

To refine the camera intrinsics in the final bundle adjustment, two key steps are required. First, the camera intrinsics must be included as parameters in the local bundle adjustment. Second, the Schur complement applied to the local hessian matrix in \cref{eq:costQuad} must extract both the extrinsic and intrinsic parameters. This increases the size of the reduced camera matrix to at least $\left(6N+3 \times 6N+3\right)$, if the camera is shared among all images and has no distortions. In the global bundle adjustment, the local hessian matrices are accumulated as in \cref{eq:globalBA} where the observations $\mathbf{x}_0$ are complemented by the input initial intrinsics. The intrisics appear thus as the observed unknowns in our \textit{pointless}  BA. Note that the local BA with camera intrinsics as unknowns should \textit{free} the intrinsics only in the very last iteration in which the hessian matrix $h$ should not be inverted. This is for two reasons: (i) inverting the hessian of a 3-image block with unknown intrinsics would be very unstable and (ii) for shared camera intrinsics it violates the sharing property.  
%
\paragraph{Limitations.} 
For highly overlapping acquisitions, such as the video acquisition of ETH3D, viewgraph pre-selection is necessary and can be done for instance through sketonization techniques~\cite{snavely2008skeletal}. Running our method on a full graph consisting of all possible relative relationships incurs a computational cost equal or higher to that of the standard BA. The same limitation and the necessity to reduce the viewgraph would apply to crowed-sourced image collections. Note that randomly reducing the number of triplets by a factor of 10 (see \cref{tab:results_all}(b) and (c)), had only a minimal impact on the reprojection error in our hessian-based BA.

\section{Conclusion}
We have presented a \textit{Pointless} Global Bundle Adjustment -- a new way to optimise camera poses which disengages explicit feature points from the adjustment. Instead, our BA implicitly incorporates the feature points through rigorous propagation of the camera hessians defined in their relative frame into the global frame.

By examining the feature reprojection errors, trajectory drift and a runtime proxy metric, we demonstrated that our bundle adjustment remains as efficient as the \sota motion averaging bundle adjustment while being competetive with traditional point-based bundle adjustments in terms of precision. 

We have presented our method as an efficient approach to the final global bundle adjustment. However, we think of \textit{pointless} BA  as more generic, and we argue that it can be integrated as an intermediary adjustment routine within any \sfm pipeline.

{\small
\bibliographystyle{ieee_fullname}
\bibliography{cvpr_prophess}
}

\end{document}